\documentclass{article}

\PassOptionsToPackage{numbers,sort&compress}{natbib}
\usepackage[preprint]{neurips_2025}

\usepackage[utf8]{inputenc}
\usepackage[T1]{fontenc}
\usepackage{hyperref}
\usepackage{booktabs}
\usepackage{amsfonts}
\usepackage{nicefrac}
\usepackage{microtype}
\usepackage{xcolor}
\usepackage{tcolorbox}

\title{Preliminary Use of Vision Language Model Driven Extraction of Mouse Behavior Towards Understanding Fear Expression}

\author{
  Paimon Goulart$^{*}$ \\
  University of California, Riverside \\
  \texttt{pgoul002@ucr.edu} \\
  \And
  Jordan Steinhauser$^{*}$ \\
  University of California, Riverside \\
  \texttt{jstei007@ucr.edu} \\
  \And
  Kylene Shuler \\
  University of California, Riverside \\
  \texttt{kshul004@ucr.edu} \\
  \And
  Edward Korzus \\
  University of California, Riverside \\
  \texttt{edkorzus@ucr.edu} \\
  \And
  Jia Chen \\
  University of California, Riverside \\
  \texttt{jiac@ucr.edu} \\
  \And
  Evangelos E.~Papalexakis \\
  University of California, Riverside \\
  \texttt{epapalex@cs.ucr.edu}
}

\begin{document}

\maketitle
\renewcommand{\thefootnote}{\fnsymbol{footnote}}
\footnotetext[1]{Equal contribution.}
\renewcommand{\thefootnote}{\arabic{footnote}}

\begin{abstract}
Integration of diverse data will be a pivotal step towards improving scientific explorations in many disciplines. This work establishes a vision-language model (VLM) that encodes videos with text input in order to classify various behaviors of a mouse existing in and engaging with their environment. Importantly, this model produces a behavioral vector over time for each subject and for each session the subject undergoes. The output is a valuable dataset that few programs are able to produce with as high accuracy and with minimal user input. Specifically, we use the open-source Qwen2.5-VL model and enhance its performance through prompts, in-context learning (ICL) with labeled examples, and frame-level preprocessing. We found that each of these methods contributes to improved classification, and that combining them results in strong F1 scores across all behaviors, including rare classes like freezing and fleeing, without any model fine-tuning. Overall, this model will support interdisciplinary researchers studying mouse behavior by enabling them to integrate diverse behavioral features, measured across multiple time points and environments, into a comprehensive dataset that can address complex research questions.
\end{abstract}

\section{Introduction}

Understanding how mice express fear and how neuronal networks encode threat and safety cues requires precise behavioral annotations synchronized with neural activity. Existing tools often lack the flexibility or scalability needed to capture the nuanced repertoire of behaviors that emerge during fear and safety learning. In this work, we propose the use of vision-language models (VLMs) as a scalable and generalizable solution to automatically annotate behavior from video, offering a powerful advancement for fields like behavioral neuroscience.

In our experimental setup, mice were conditioned to associate a specific environment with threat and then tested in either the threatening or a safe context. Behavior during these sessions was recorded, and the goal was to annotate the mouse’s actions at every second of video, producing a rich behavior vector that can be integrated with other neural datasets (e.g., calcium imaging, Neuropixels). Moving beyond the traditional binary measure of fear as freezing, our framework allows for the classification of a wide range of behaviors, from exploratory sniffing to active fleeing, enabling more nuanced interpretation of safety learning and threat perception.

While existing tools such as DeepLabCut and MoSeq have laid the foundation for video based behavior analysis, they come with notable limitations \cite{nath2019deeplabcut, lin2024motionseq}. DeepLabCut requires extensive user input and may fail to capture subtle exploratory behaviors. MoSeq minimizes user input but is limited to clustering high-frequency stereotyped motifs. In contrast, our VLM approach supports direct behavior labeling via natural language prompts and offers both flexibility and interpretability, with minimal setup or retraining.

Our model takes as input a video of mouse behavior and, given a user defined prompt, returns per-second annotations of behavior and environment. These annotations can then be used independently or integrated with other experiment modalities to ask more complex scientific questions about cognition, emotion, and learning. Our code is available at \url{https://github.com/Pie115/VLM-Labeling}.
\section{Methods}

\subsection{Data Description}
The goal of our model is to extract key indicators of mouse behavior during fear expression and safety learning. Mice were trained to associate a specific environment with threat, and then tested in either the same threatening environment or a similar but safe one. The model annotates these video recordings to identify behaviors that reflect fear or safety, depending on the environment.

Fear-related behaviors are defensive and typically reflect a high perception of threat. These include freezing, fleeing, and periphery bias. Freezing refers to the complete absence of movement other than breathing and is a canonical rodent fear response used to avoid detection \cite{Blanchard_2001, Takemoto_Song_2019, Gerlai_1998, Laxmi_Stork_Pape_2003, Clark_Drummond_Hoyer_Jacobson_2019, Roelofs_Dayan_2022, Laughlin_Moloney_Samels_Sears_Cain_2020, Trott_Hoffman_Zhuravka_Fanselow_2022, Smith_Markowitz_Hoffman_Fanselow_2022}. Fleeing involves sudden darting or escape behavior in response to a perceived threat \cite{Blanchard_2001, Griebel_Belzung_Misslin_Vogel_1993, Gerlai_1998}. Periphery bias refers to the tendency of mice to remain near the chamber walls; reduced center exploration is a common proxy for anxiety or fear \cite{Romeo_Mueller_Sisti_Ogawa_McEwen_Brake_2003}.

In contrast, safety-related behaviors suggest a perception of low threat. These include grooming, a stereotyped self-cleaning behavior seen in relaxed states \cite{Gerlai_1998, Naik_Ma_Munyeshyaka_Leibenluft_Li_2024}, and exploratory behavior, in which the animal actively investigates its surroundings through nose-poking, rearing, and sniffing \cite{Griebel_Belzung_Misslin_Vogel_1993, Justice_Carter_Beck_Gioscia-Ryan_McQueen_Enoka_Seals_2013, Laxmi_Stork_Pape_2003}. Rearing refers to when the mouse lifts its forepaws and stands on its hindlimbs, while sniffing involves head movement and an alert upright posture indicative of curiosity. Center exploration is also considered a safety signal, as mice avoid the center of an open arena under threat.

In total, our dataset consists of 3240 seconds of annotated mouse behavior across multiple video sessions. After merging grooming and exploring into a single category representing safety behavior, the label distribution is as follows: freezing (410/3240, 12.7\%), fleeing (21/3240, 0.6\%), and grooming/exploring (2809/3240, 86.7\%).

\subsection{Vision Language Model}

In order to consistently and accurately annotate the behavior data of mice, we opt towards a vision language model (VLM). Specifically, we use Qwen2.5-VL as it is an open source model that achieves state-of-the-art performance \cite{Qwen-VL, Qwen2-VL, Qwen2.5-VL}. Although this model is pre-trained on a broad mixture of text and vision data, we discovered it is not well tuned for specific scientific or animal behavior tasks. However, despite this limitation, we find that it is still possible to repurpose through careful prompt engineering and strategic input formatting. Notably, we do not fine-tune the model weights at any stage; all task-specific behavior recognition is achieved through prompting and in-context learning alone.

In our setup, Qwen2.5-VL receives either a full video or individual frames as its visual input, paired with a textual prompt instructing the model to identify the behavior observed.

\subsection{Video Preprocessing}

To obtain a per-second behavior label vector aligned with each video, we encountered several practical challenges. Initially, we found it extremely difficult for the model to accurately estimate the number of seconds, let alone generate a behavior prediction for each one, when given the full video as input. In general, VLMs are known to struggle with temporal reasoning, especially in tasks such as second-by-second and timelapse interpretation \cite{imam2025multimodalllmsvisualtemporal}. While recent work has begun to address these limitations, models like Qwen2.5-VL still exhibit a noticeable lack of temporal understanding in long video inputs \cite{Qwen2.5-VL}.

To address this, we split each video into individual one-second segments and process them separately. Rather than asking the model to reason over the full duration of the video and identify the behavior occurring at each second, we provide the model with a single second of video at a time and ask it to classify the dominant behavior within that interval. This approach reduces the burden of temporal reasoning and allows the model to focus solely on visual understanding. Importantly, by restricting the model to output only one label at a time, we also reduce the likelihood of hallucinations or cascading errors that often arise in more open ended generations \cite{kalai2025languagemodelshallucinate}. As a result, our method yields behavior predictions that are not only more accurate but also directly aligned with the desired temporal resolution, one label per second of video.

Another limitation we identified stems from how Qwen2.5-VL processes videos. Since the model samples only a subset of frames from the input video, it is entirely possible for it to miss minor actions or subtle movements that are critical for accurately identifying mouse behavior \cite{Qwen2.5-VL, li2025improving}. To address this, we preemptively split each video into its individual frames and feed each frame into the model independently. By doing so, we effectively force the model to observe the full content of the video, which we hypothesize helps it better distinguish between behaviors, especially those triggered by brief or subtle frame-to-frame transitions.

\subsection{In-Context Learning}

In-Context Learning (ICL) is a technique that enables language models to perform classification, regression, and other tasks without the need for fine-tuning or task-specific training \cite{min2022rethinkingroledemonstrationsmakes, coda2023meta}. While ICL has traditionally been explored in text only settings, recent work has extended its use to visual and multimodal tasks as well \cite{ferber2024incontextlearningenablesmultimodal, kim2024videoiclconfidencebasediterativeincontext, zhang2025videoincontextlearningautoregressive}. Despite this progress, the application of ICL within VLMs, especially for structured video understanding in scientific domains, remains relatively underexplored.

Given these gaps, we hypothesize that ICL can be particularly effective in helping VLMs reason about complex and domain specific behaviors, such as those observed in laboratory mice, by providing a few labeled visual examples as context. In our setup, we explore the use of ICL to improve behavior labeling by supplying the model with a small number of annotated image label pairs (few-shot demonstrations) directly in the prompt. The goal is to improve recognition of subtle or ambiguous behaviors without requiring additional training or domain adaptation.

\subsection{Labeling Pipeline}

\begin{figure}[h]
    \centering
    \includegraphics[width=0.55\linewidth]{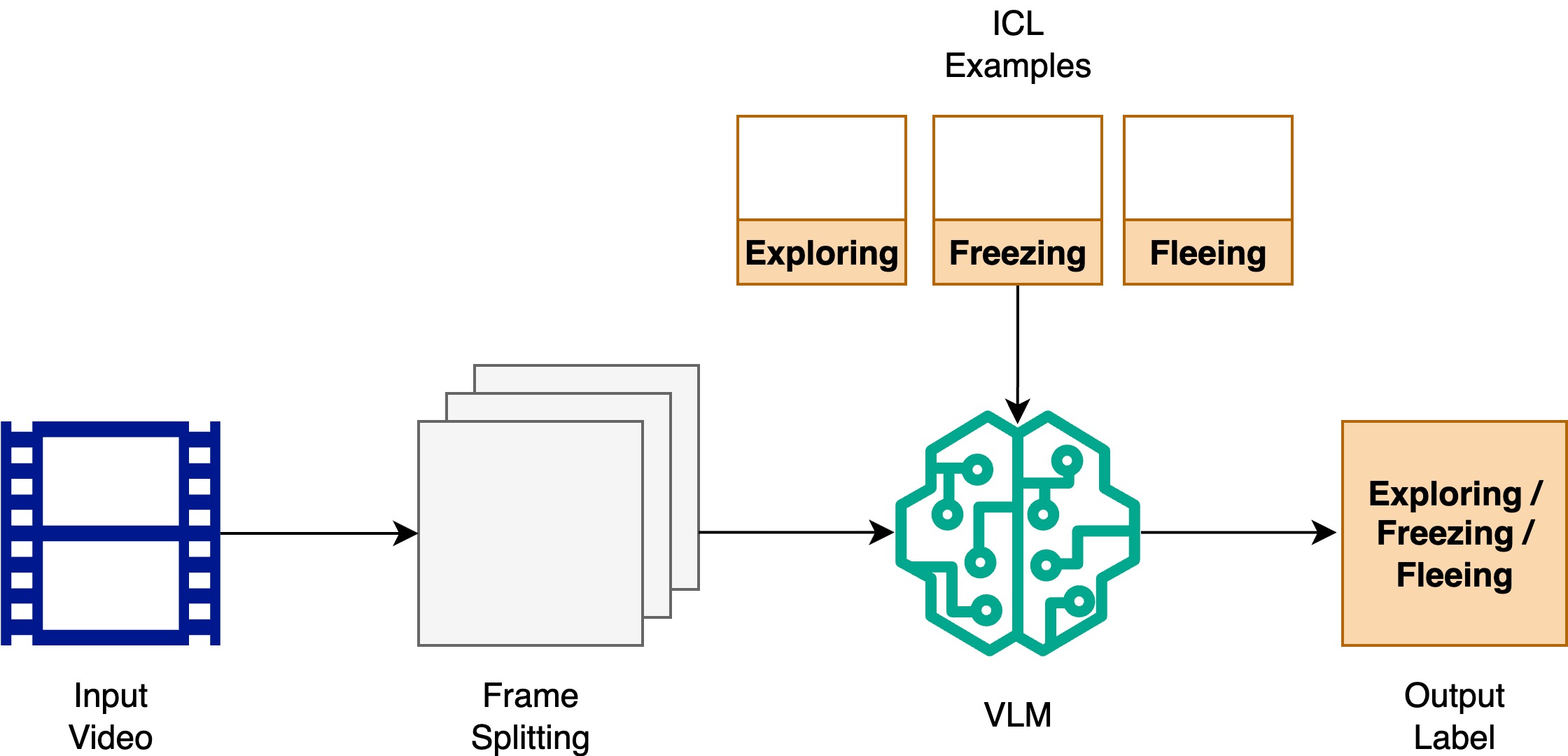}
    \caption{
        Overview of our video labeling pipeline.
    }
    \label{fig:vlm_pipeline}
\end{figure}

By combining all these methods we come up with the following labeling pipeline shown in \ref{fig:vlm_pipeline}. First, each input video is split into one-second segments. Each  segment is then decomposed into its individual frames, which are fed into the VLM as a single visual unit. This input ensures that the model observes all visual content within the second, rather than relying on its internal frame sampling, which could miss important frames.

\begin{figure}[h]
    \centering
    \begin{tcolorbox}[colframe=blue!50!black, colback=blue!5!white, title=Simple Prompt with In-Context Learning]
    \textbf{Task:}  

    Label the mouse's behavior seen per second for this n-second video segment such that we have a label for each second of the video. For example: \texttt{[behavior\_1, behavior\_2, ..., behavior\_n]}.  
    Annotation key: \texttt{Freezing}, \texttt{Fleeing}, \texttt{Exploring/Grooming}.  
    Only output the vector! Given the following examples, label the last video to the best of your ability.

    \vspace{0.3cm}
    \textbf{Examples:}

    \textbf{Video/Frames 1:} \includegraphics[height=1.5cm]{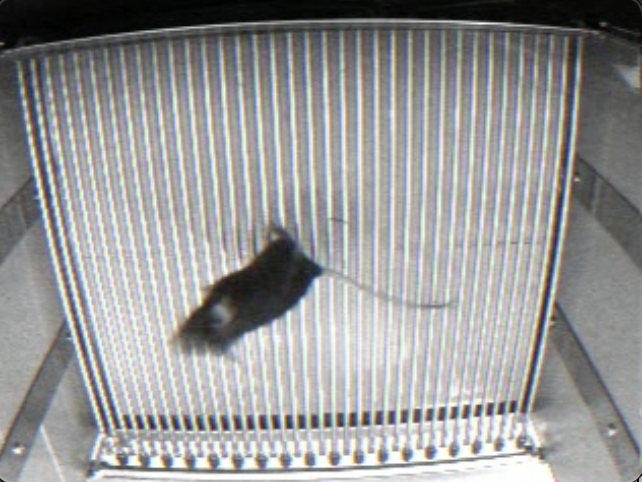} $\rightarrow$ \texttt{[Exploring/Grooming]}

    \textbf{Video/Frames 2:} \includegraphics[height=1.5cm]{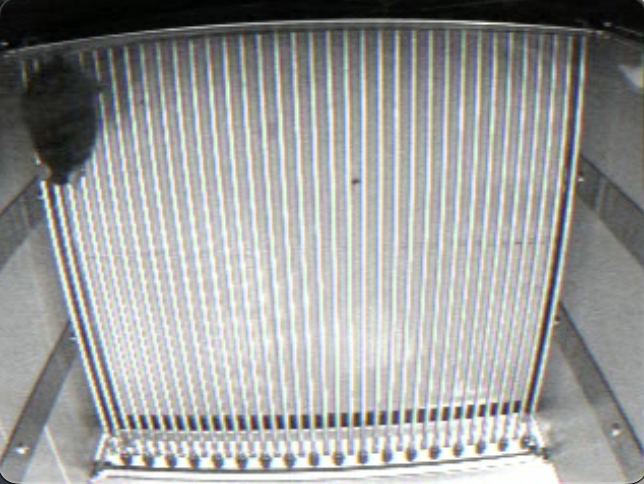} $\rightarrow$ \texttt{[Freezing]}

    \textbf{Video/Frames 3:} \includegraphics[height=1.5cm]{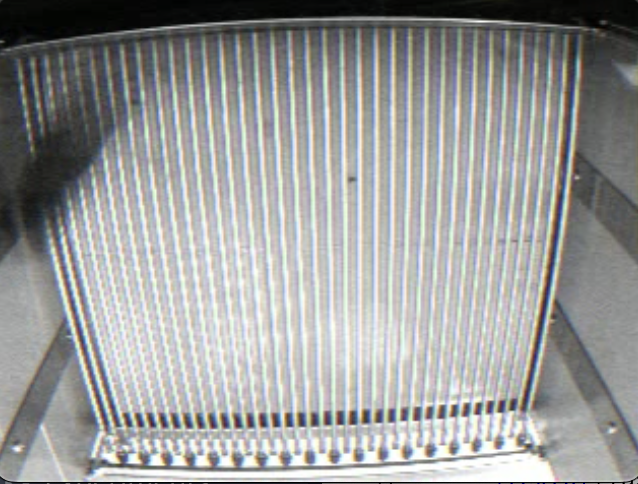} $\rightarrow$ \texttt{[Fleeing]}

    \vspace{0.3cm}
    \textbf{Target (Video/Frames):} 
    \fbox{\parbox[c][1.5cm][c]{2cm}{\centering \textbf{Input}}} 
    $\rightarrow$ \texttt{[ \ \ ]}
    \end{tcolorbox}
    \caption{Simple ICL prompt: the model receives a few labeled examples and a target video segment, and is asked to choose one behavior label.}
    \label{fig:icl_simple_prompt}

\end{figure}

\begin{figure}[h]
    \centering
    \begin{tcolorbox}[colframe=blue!50!black, colback=blue!5!white, title=Complex Prompt with In-Context Learning]
    \textbf{Task:}  

    Label the mouse’s behavior in this n-second clip.  
    Return exactly n label(s) in a Python list \texttt{[l1, ..., l1]}.

    Allowed: \texttt{Freezing}, \texttt{Fleeing}, \texttt{Exploring/Grooming}.  
    \texttt{Freezing} = absolutely no visible movement across the whole second (no head/ear/whisker/tail or body motion).  
    \texttt{Exploring/Grooming} = any visible movement that isn’t fast fleeing; includes slow stepping in place, head/whisker/ear/tail motion, sniffing, re-orienting, or brief rearing with little displacement.  
    \texttt{Fleeing} = fast, sustained locomotion, large displacement, motion blur, or dashing.  

    If unsure between fleeing and exploring, choose fleeing if movement is more rapid.  
    Rules: if any movement is seen in any frames, do NOT output \texttt{Freezing}.  
    Output only the list.

    \vspace{0.3cm}
    \textbf{Examples:}

    \textbf{Video/Frames 1:} \includegraphics[height=1.5cm]{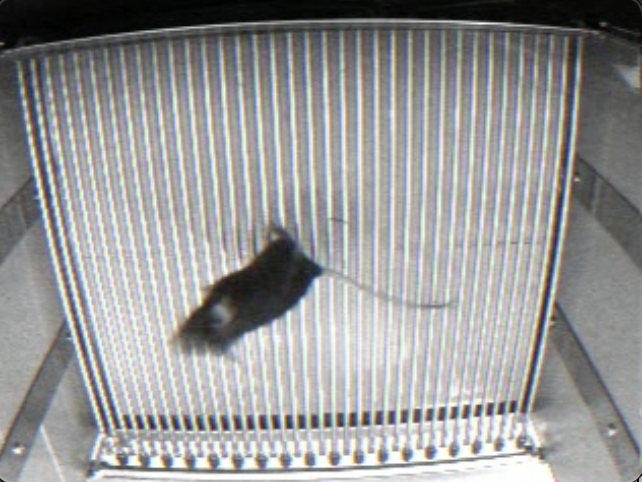} $\rightarrow$ \texttt{[Exploring/Grooming]}

    \textbf{Video/Frames 2:} \includegraphics[height=1.5cm]{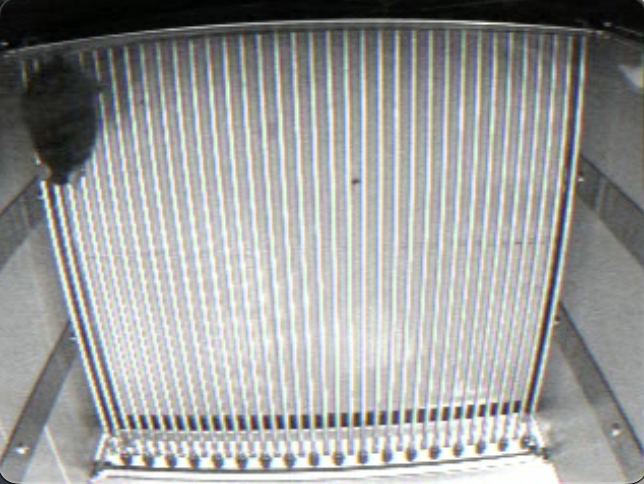} $\rightarrow$ \texttt{[Freezing]}

    \textbf{Video/Frames 3:} \includegraphics[height=1.5cm]{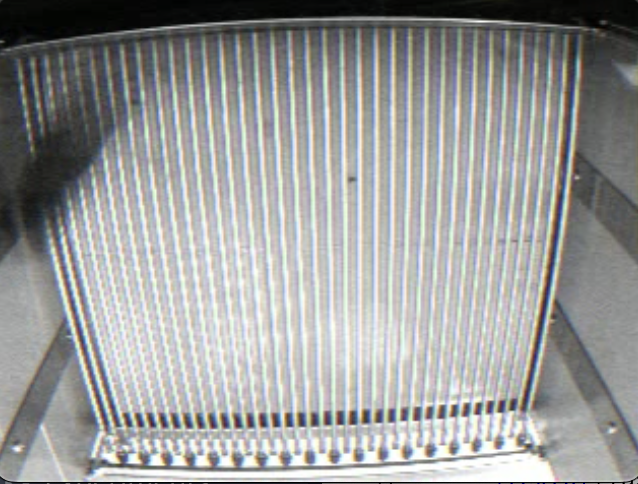} $\rightarrow$ \texttt{[Fleeing]}

    \vspace{0.3cm}
    \textbf{Target (Video/Frames):} 
    \fbox{\parbox[c][1.5cm][c]{2cm}{\centering \textbf{Input}}} 
    $\rightarrow$ \texttt{[ \ \ ]}

    \end{tcolorbox}
    \caption{Complex ICL prompt: the model is given definitions for each behavior alongside example demonstrations. Note that non-ICL uses the same prompt without examples.}
    \label{fig:icl_complex_prompt}
\end{figure}

For each input video, we prepend a small number of labeled examples drawn from a held out set of videos. These ICL examples are formatted in the same way; being one-second segments decomposed into individual frames and paired with their known behavior labels. All examples and the target video segment are then fed into the model using one of two prompts; a simple prompt or a complex prompt. The simple prompt asks the model to choose one of the behavior labels, while the complex prompt includes behavior definitions to provide additional guidance. An example of the simple and complex prompt is shown in Figures \ref{fig:icl_simple_prompt} and \ref{fig:icl_complex_prompt}. The simple prompt follows the same format, but instead omits the behavior definitions and is asked to choose one of the labels directly. 

Finally, the model returns a label for the given video segment. This process is repeated for each second of video, resulting in a complete per-second behavior vector aligned with the original video length.

\section{Experiments}
To evaluate the contribution of each component in our labeling pipeline, we conducted a series of experiments assessing their individual and cumulative effects on classification performance. We began with the off-the-shelf Qwen2.5-VL model using both the simple and complex prompts without any ICL or preprocessing. We then incrementally incorporated ICL, followed by the full frame-splitting strategy. This allows us to isolate the impact of each method and quantify their combined effect on performance.

We evaluated each configuration on all 3240 seconds of data using per-class F1 scores to account for the heavy class imbalance.

\begin{figure}[!ht]
    \centering
    \includegraphics[width=\linewidth]{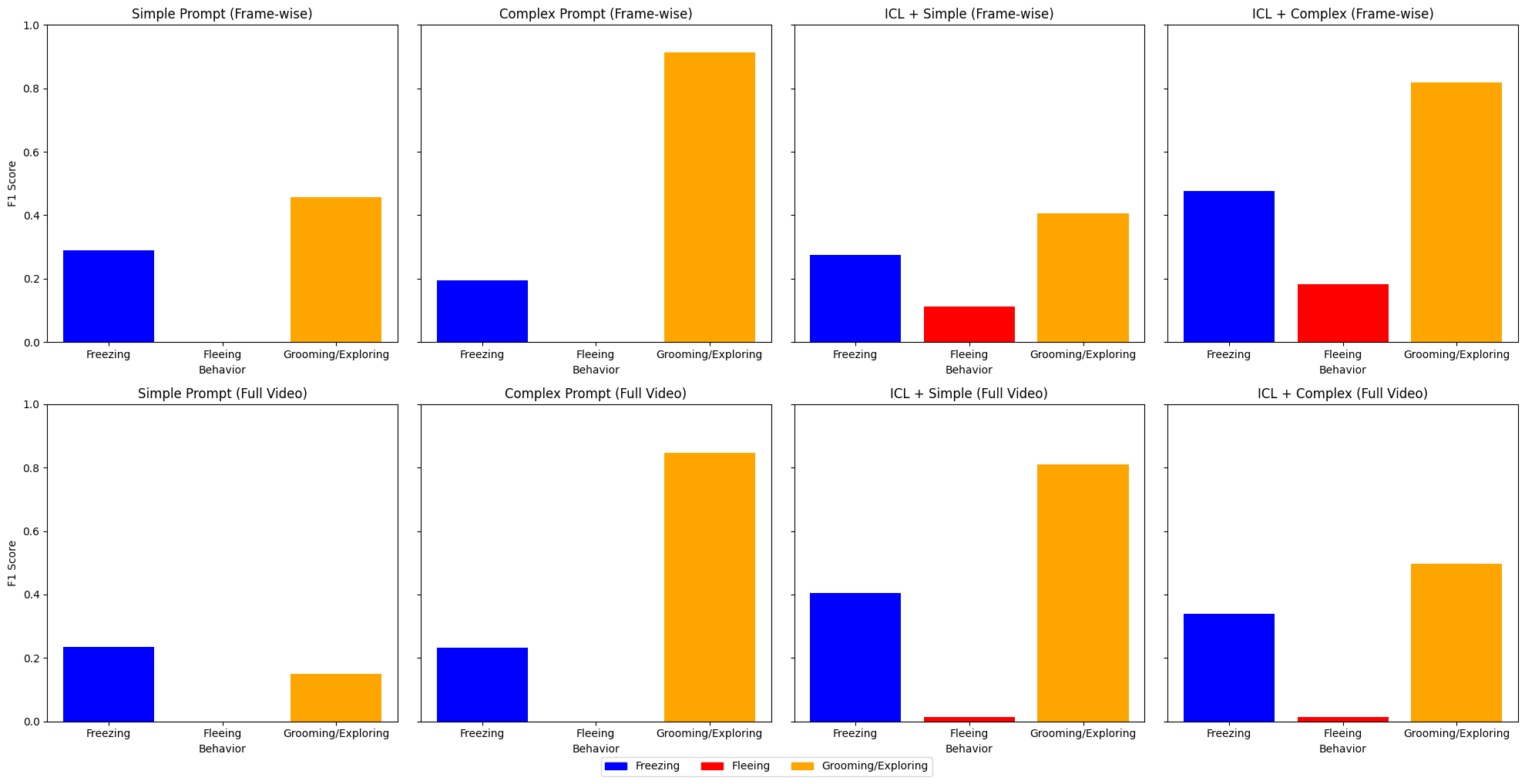}
    \caption{Per class F1 scores across all pipeline configurations. We compare the effects of using simple vs. complex prompts, with and without ICL, as well as the impact of full video input vs. frame-wise input. ICL and frame splitting significantly improve performance, especially for rare behaviors.}
    \label{fig:f1_comparisons}
\end{figure}

We see the results of these configurations in Figure \ref{fig:f1_comparisons}. Across all configurations, we observe a clear progression in model performance as each component of the pipeline is added. Starting with the off-the-shelf model, we find that the simple prompt achieves moderate F1 scores for the dominant class (grooming/exploring), but performs poorly on freezing and fails entirely to capture fleeing behavior. The complex prompt improves Grooming/Exploring and slightly boosts freezing recognition, but still misses fleeing.

The introduction of ICL provides notable gains, especially for the rare classes. With ICL, the model begins to label fleeing behavior and achieves consistent improvements in freezing detection across both prompt styles. This suggests that visual demonstrations help guide the model toward rare or ambiguous behaviors that are hard to detect from the raw video alone.

Finally, incorporating full frame-splitting further amplifies these improvements. The combination of ICL and frame-wise processing (top right plot) yields the best overall results, with strong F1 scores across all behaviors. Notably, ICL + Complex Prompt with frame-wise input recovers both freezing and fleeing with measurable accuracy.

These findings confirm that prompt design, ICL, and preprocessing each contribute meaningfully, with their combination yielding the strongest results.

\section{Conclusions}

We present a VLM pipeline for annotating mouse behavior with minimal supervision, enabling second-by-second labeling of fear and safety related actions. Through experimentation, we show that prompt design, ICL, and frame-level preprocessing each contribute significantly to performance, with their combination yielding the most accurate results, especially for rare but scientifically critical behaviors like freezing and fleeing.

This work demonstrates the feasibility of using VLMs to generate behavioral annotation vectors that can be integrated with neural recordings or other experimental data, ultimately leading to insights into threat and safety learning, among other animal behaviors and cognitions.

Future work will explore incorporating temporal context across video segments, labeling of longer segments, richer environmental annotations, and human-in-the-loop correction mechanisms to further enhance label quality.

{
    \small
    \bibliographystyle{unsrtnat}
    \bibliography{main}
}

\end{document}